\documentclass[letterpaper, 10 pt, conference]{ieeeconf}  % Comment this line out if you need a4paper

\IEEEoverridecommandlockouts                              % This command is only needed if 
\usepackage{amsmath} % assumes amsmath package installed

\usepackage{amssymb}  % assumes amsmath package installed
\usepackage{amsthm}
\usepackage{bm}
\usepackage{todonotes}
\usepackage{multicol}
\usepackage[linesnumbered,ruled,noend]{algorithm2e}
\usepackage{xcolor}

\SetCommentSty{mycommfont}

\makeatletter
\newcommand{\removelatexerror}{\let\@latex@error\@gobble}
\makeatother

\newcommand{\X}{\mathcal{X}}
\newcommand{\W}{\mathcal{W}}
\newcommand{\U}{\mathcal{U}}
\newcommand{\Y}{\mathcal{Y}}
\newcommand{\K}{\mathcal{Y}_k}
\newcommand{\Z}{\mathcal{Z}}
\newcommand{\A}{\mathcal{A}}

\newcommand{\nx}{n_x}
\newcommand{\nc}{n_u}
\newcommand{\ny}{n_y}
\newcommand{\nk}{n_k}
\newcommand{\nz}{n_z}
\newcommand{\aug}{a}

\title{\LARGE \bf
Synthesizing Stable Reduced-Order Visuomotor Policies for \\ Nonlinear Systems via Sums-of-Squares Optimization
}

\author{Glen Chou and Russ Tedrake% <-this % stops a space
\thanks{Computer Science and Artificial Intelligence Laboratory, Massachusetts Institute of Technology, Cambridge, MA 02139.
        {\tt\small \{gchou, russt\}@mit.edu}. This work is supported in part by the MIT Quest for Intelligence, and Amazon.com Services LLC Award \#2D-06310236.}%
}

\begin{document}

\maketitle
\thispagestyle{empty}
\pagestyle{empty}

%%%%%%%%%%%%%%%%%%%%%%%%%%%%%%%%%%%%%%%%%%%%%%%%%%%%%%%%%%%%%%%%%%%%%%%%%%%%%%%%
\begin{abstract}

We present a method for synthesizing dynamic, reduced-order output-feedback polynomial control policies for control-affine nonlinear systems which guarantees runtime stability to a goal state, when using visual observations and a learned perception module in the feedback control loop. We leverage Lyapunov analysis to formulate the problem of synthesizing such policies. This problem is nonconvex in the policy parameters and the Lyapunov function that is used to prove the stability of the policy. To solve this problem approximately, we propose two approaches: the first solves a sequence of sum-of-squares optimization problems to iteratively improve a policy which is provably-stable by construction, while the second directly performs gradient-based optimization on the parameters of the polynomial policy, and its closed-loop stability is verified \textit{a posteriori}. We extend our approach to provide stability guarantees in the presence of observation noise, which realistically arises due to errors in the learned perception module. We evaluate our approach on several underactuated nonlinear systems, including pendula and quadrotors, showing that our guarantees translate to empirical stability when controlling these systems from images, while baseline approaches can fail to reliably stabilize the system.

\end{abstract}

%%%%%%%%%%%%%%%%%%%%%%%%%%%%%%%%%%%%%%%%%%%%%%%%%%%%%%%%%%%%%%%%%%%%%%%%%%%%%%%%
\section{Introduction}

For autonomous robots to be effective in the real world, we need provable assurances on their safety and reliability. In these unstructured settings, robots typically lack full state information, and must complete tasks while controlling using only sensor measurements (i.e., outputs); this perception-based control problem is known as \textit{output-feedback}. For robots of interest with nonlinear dynamics, synthesizing output-feedback controllers is known to be difficult. Partial observability and sensor noise require that output-feedback controllers extract information from a \textit{history} of observations to stabilize the system. Moreover, in many domains of interest, these observations are non-smooth and high-dimensional (e.g., images), which can increase the amount of data needed to learn a good visuomotor control policy, i.e., a policy which takes (a history of) images as input and returns a control action. Finally, underactuation due to input limits and nonlinearities in the dynamics can lead to numerous local minima when attempting to optimize output-feedback control policies. To address these difficulties, recent work (e.g., \cite{DBLP:journals/jmlr/LevineFDA16, DBLP:journals/corr/SchulmanWDRK17}) leverages the power of neural networks (NNs) and deep reinforcement learning (RL) to tackle the output-feedback problem. However, the resulting control policies are complex, rendering the learning process data-hungry and the closed-loop stability of the system difficult to certify. 

\begin{figure}
    \centering
    \includegraphics[width=\linewidth]{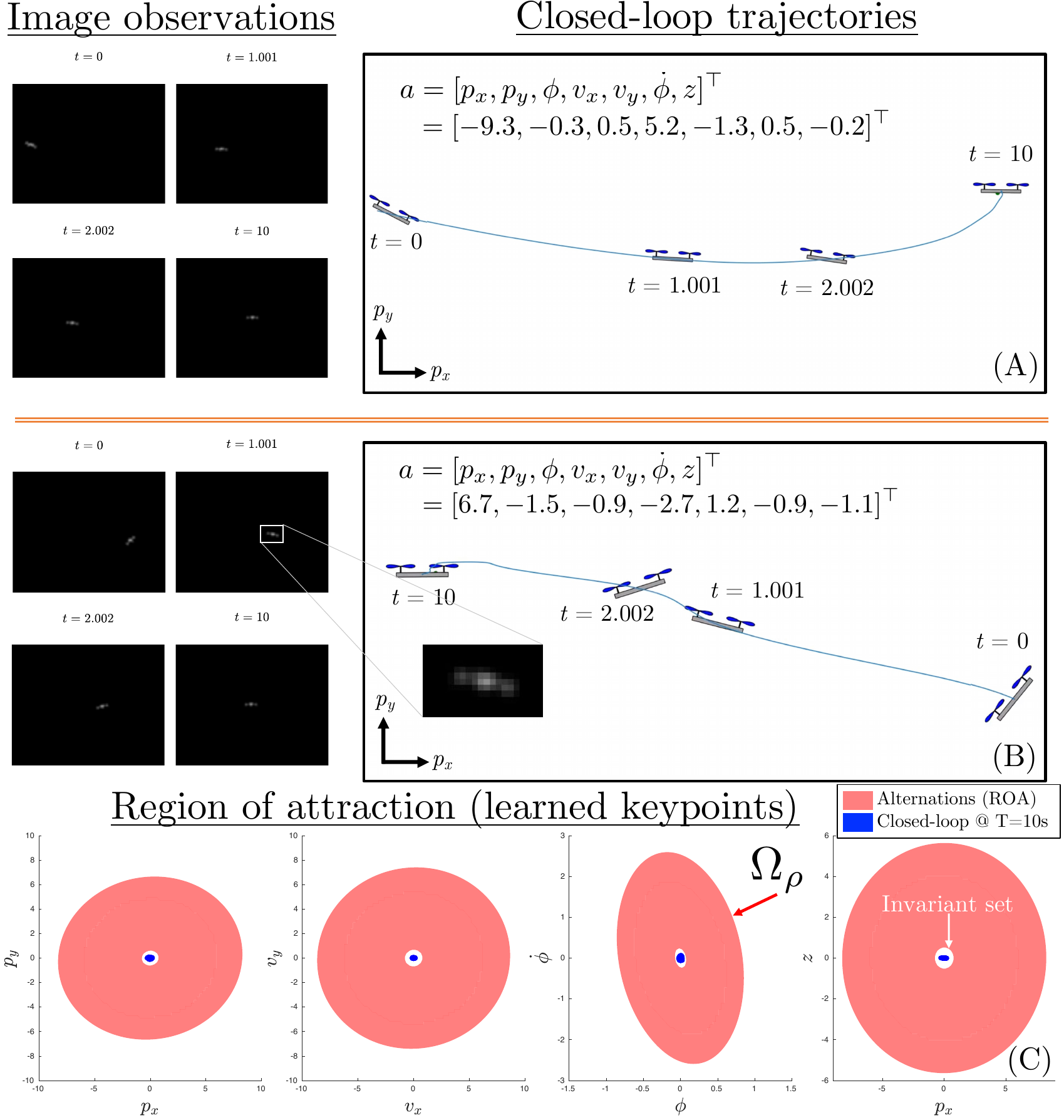}\vspace{-10pt}
    \caption{In this paper, we synthesize dynamic-output-feedback controllers that stabilize nonlinear systems from pixels. \textbf{(A-B)}: Stabilizing a planar quadrotor to the origin, with a learned perception map $\hat h_e$ in the loop, from initial conditions (ICs) $\aug = [p_x, p_y, \phi, v_x, v_y, \dot \phi, z]^\top$. Here, $z \in \mathbb{R}$ is the controller latent variable. Left: Grayscale 128x96 images input to the controller. Right: Time-lapse of stabilized trajectories. \textbf{(C)}: Slices (all other states set to zero) of a certified inner approximation of the closed-loop region of attraction (ROA) in red. While the system may not converge exactly to the origin due to errors in the learned perception map, all states in the ROA are guaranteed to converge to an invariant set (in white) around the origin. To empirically show the set's invariance, we plot (in blue) the states reached after $10$s have elapsed, for 500 ICs sampled from the ROA.\vspace{-7pt} }
    \label{fig:pvtol_rollout}
    \vspace{-16pt}
\end{figure}

In this paper, we challenge the notion that complex controllers are required to stabilize nonlinear robotic systems from high-dimensional sensor inputs like images. In particular, we show that given a useful reduction of the high-dimensional observations which can be learned from data (such as keypoints rigidly attached to the robot), simple \textit{dynamic} (i.e., stateful) policies, which are linear or a low-degree polynomial in the reduced observations, can effectively stabilize a variety of nonlinear, underactuated systems. Due to the simplicity of these controllers, our method can leverage powerful tools like sums-of-squares (SOS) programming to synthesize provably-stable output-feedback policies. Inspired by the scalability of gradient-based policy learning in RL \cite{DBLP:conf/iclr/0028MNRMGM22}, we also provide a method that directly learns the parameters of the polynomial controller through gradient descent (GD) (Sec. \ref{sec:method_gradient}). Due to this parameterization, a region of attraction (ROA) for the learned controller can be readily verified in a post-hoc fashion by solving a set of smaller SOS optimizations. Finally, a key strength of our controllers is that they \textit{do not explicitly reconstruct the full state}. This is critical as it 1) reduces the problem dimension, keeping the SOS problems tractable for high-dimensional systems, and 2) sidesteps the need for an accurate initial state estimate. Our specific contributions are:
\begin{itemize}
    \item Two approaches for synthesizing dynamic reduced-order output-feedback controllers (via SOS programming and via gradient-based optimization)
    \item An extension for verifying robust closed-loop stability under observation error, enabling end-to-end stability guarantees when using images as input and an imperfect learned perception module in the control loop
    \item Validation on a variety of nonlinear systems, showing that our simple polynomial controllers match or outperform baselines in optimizing visuomotor policies
\end{itemize}

\section{Related Work}

Many learning-based approaches for designing control policies for nonlinear robotic systems from rich sensor inputs like images have been recently proposed \cite{DBLP:journals/jmlr/LevineFDA16, DBLP:journals/corr/SchulmanWDRK17, DBLP:conf/aistats/BanijamaliSGB018}. These approaches typically rely on model-based or model-free reinforcement learning, which leverage data to learn complex NN controllers which take images as input. However, these controllers are complex, difficult to analyze, and may not reliably stabilize the system. On the other hand, many model-based methods exist for synthesizing stable state-feedback controllers for nonlinear systems, and we know that simple, low-degree polynomial controllers can stabilize many robots. Using tools like SOS optimization \cite{10.5555/2430708}, Lyapunov analysis \cite{DBLP:conf/icra/MajumdarAT13}, contraction \cite{arxiv_version}, and barriers \cite{DBLP:conf/cdc/Clark21} can be used to algorithmically synthesize stable polynomial controllers. 

For the specific case of linear systems, techniques like linear-quadratic-Gaussian (LQG) control \cite{astrom} and its reduced-order counterpart \cite{doi:10.2514/6.1984-1035} can efficiently synthesize dynamic-output-feedback controllers. While there is extensive theoretical study on sufficient conditions for nonlinear output-feedback stabilization \cite{Isidori1992DisturbanceAA, DBLP:journals/tac/AstolfiC02}, there are far fewer methods for algorithmic synthesis of such controllers. Most existing work leverages SOS optimization, and focuses on finding static output-feedback controllers \cite{DBLP:conf/cdc/Baldi16, 5975729} (i.e., output-feedback controllers which only take in the current observation as input) or decouples the state-feedback and full-order observer design problems \cite{Tan2006NonlinearCA, Zheng2008NonlinearOF, DBLP:conf/aucc/ManchesterS14}. While in principle, dynamic controllers can be written as static controllers by augmenting the observations with the latent variables, directly applying these methods when the latent dynamics are also being optimized, as in our method, leads to additional nonconvexities. Moreover, these methods are evaluated on low-dimensional systems, and are driven by low-dimensional observations. In contrast, we solve a reduced-order dynamic-output-feedback policy synthesis problem, and apply it on high-dimensional robotic systems for image-based control.

Finally, there is recent work in Lyapunov-based control from high-dimensional observations. Learned approximate Lyapunov functions are used in \cite{DBLP:journals/ral/DawsonLGF22} to stabilize from LiDAR; however, the Lyapunov conditions are not actually certified. Other work uses images with barriers \cite{DBLP:conf/corl/DeanTCRA20}; however, the full state must be directly invertible from a single observation, which is not possible for most robotic systems. More recent work \cite{DBLP:conf/wafr/ChouOB22} leverages contraction theory to control safely from images, but does so by decoupling the state-feedback and estimation problems and requires estimation of the full state. 

\section{Preliminaries}\label{sec:prelim}
\begin{figure}
    \centering
    \includegraphics[width=\linewidth]{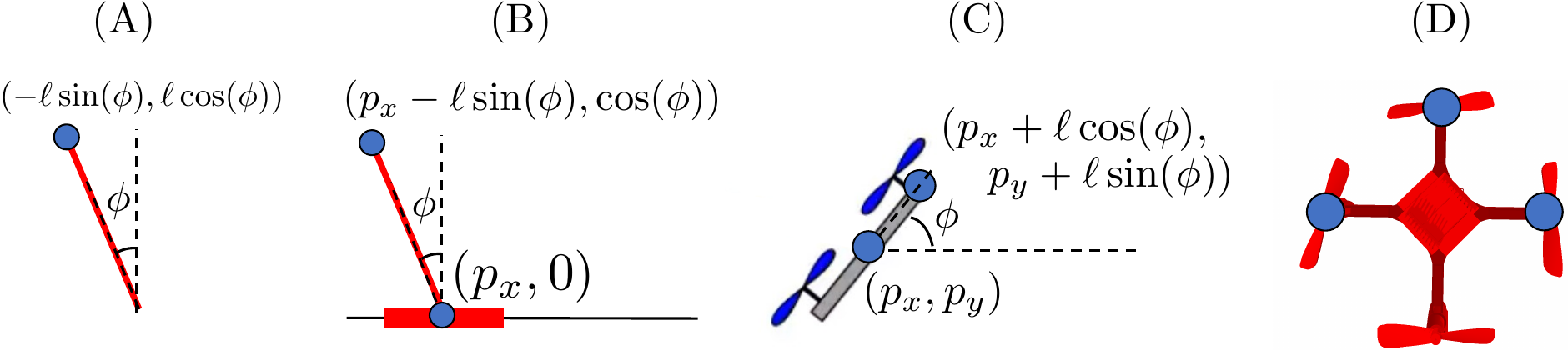}\vspace{-10pt}
    \caption{Keypoints (blue) used for (\textbf{A}): pendulum, ($y_k^* \in \mathbb{R}^2$) (\textbf{B}): cart-pole ($y_k^* \in \mathbb{R}^4$), (\textbf{C}): 2D quadrotor ($y_k^* \in \mathbb{R}^4$), (\textbf{D}): 3D quadrotor ($y_k^* \in \mathbb{R}^9$).}
    \label{fig:keypoints}
    \vspace{-16pt}
\end{figure}

We consider control-affine, partially-observed nonlinear systems, with state space $\X \subseteq \mathbb{R}^{\nx}$, control space $\U \subseteq \mathbb{R}^{\nc}$, and observation space $\Y \subseteq \mathbb{R}^{\ny}$,
\begin{subequations}\label{eq:system}
    \begin{align}
    \dot x(t) &= f(x(t),u(t)) = f_1(x(t)) + f_2(x(t)) u(t)\label{eq:dyn} \\
    y(t) &= h(x(t)),\label{eq:obs}
\end{align}
\end{subequations}
where $f: \X \times \U \rightarrow \bigcup_{x\in\X} T_x\X$ and $h: \X \rightarrow \Y$, and where $T_x \X$ is the tangent space of $\X$ at $x$. In this paper, we assume $f_1: \X \rightarrow \bigcup_{x\in\X} T_x\X$ and $f_2: \X \rightarrow \mathbb{R}^{\nx \times \nc}$ are polynomial functions of $x$, and that $\Y$ contains high-dimensional image observations. We do not assume $h$ is polynomial. Since designing controllers directly as a function of the pixels, which may be a non-smooth function of the state (due to aliasing from finite resolution images), can be difficult, we assume knowledge of an approximate smooth reduced-dimensional representation of the information in the images. Popular visual representations can be used here (e.g., dense descriptors \cite{DBLP:conf/corl/FlorenceMT18}) if they are learned such that they are a simple (roughly polynomial) function of $x$. In this paper, we use keypoints, denoted as $y_k^* \in \K \subseteq \mathbb{R}^{\nk}$, which are points in the workspace that are rigidly attached to the robot (see Fig. \ref{fig:keypoints} for examples). Keypoints are a common feature representation used in computer vision \cite{DBLP:conf/iccv/LagunaRPM19} and robotics \cite{DBLP:conf/isrr/ManuelliGFT19} for pose estimation. For robots modeled as rigid bodies, $y_k^*$ can be written as a polynomial function of the state, with a change of variables (see Sec. \ref{sec:results} for examples). An approximate map from images to keypoints $\hat h_e: \Y \rightarrow \K$, i.e., a keypoint \textit{extractor}, can be learned in a supervised fashion from a labeled dataset of images and keypoints. Here, $\hat h_e$ need not be polynomial; in this paper, we represent $\hat h_e$ as a convolutional neural network (CNN), and train it to directly output keypoints $y_k$. The resulting keypoints are
\begin{equation}\label{eq:kp}
\begin{array}{rl}
    y_k(t) = \hat h_e(h(x(t))) = & h_k(x(t)) + w(t)  \\
     \doteq &  y_k^*(t) + w(t)
\end{array}
\end{equation}
where $h_k: \X \rightarrow \K$ is polynomial in $x$ and $w \in \W \subseteq \mathbb{R}^{\nk}$ is a bounded disturbance which models the error in the keypoints predicted by the learned keypoint extractor $y_k$, relative to the perfect keypoints $y_k^*$.

\subsection{Lyapunov analysis and SOS optimization}\label{sec:prelim_sos}

For a system $\dot x = f(x)$ with equilibrium point $x_0$, if we can find a $C^1$ Lyapunov function $V: \X \rightarrow \mathbb{R}_{\ge 0}$ satisfying
\begin{subequations}\label{eq:lyap}
\begin{align}
	V(x_0) = 0,\quad \dot V(x_0) = 0, \\ 
	x \in \Omega_\rho^x \wedge x \ne x_0 \Rightarrow  V(x) > 0 \wedge \dot V(x) < 0,
\end{align}
\end{subequations}
where $\Omega_\rho^x \doteq \{x \mid V(x) \le \rho\}$ is the $\rho$-sublevel set of $V$, then $\Omega_\rho^x$ is contained in the system's region of attraction (ROA).
Finding a $V$ satisfying \eqref{eq:lyap} requires enforcing non-negativity of polynomials ($V$ and $-\dot V$) over a basic semialgebraic set (i.e., a set defined by a finite number of polynomial (in)equalities). While this is NP-hard, we can efficiently enforce that a polynomial is a sum of squares (SOS), which implies non-negativity. A polynomial $p$ of degree $d$ in indeterminate variables $x_1, \ldots, x_n$, $p(x_1, \ldots, x_n)$, is SOS if it can be written as $\sum_{i=1}^m q_i^2(x)$, where $q_i(\cdot)$ are polynomials. This is equivalent to the existence of a $Q \succeq 0$ such that $p(x) = \tilde{m}(x)^\top Q \tilde{m}(x)$, where $\tilde{m}(x)$ is a polynomial basis; $Q$ can be found with semidefinite programming.

\section{Problem Formulation}
We wish to design a dynamic-output-feedback controller $k: \K \times \Z \rightarrow \U$, with latent dynamics $l: \Z \times \K \rightarrow \bigcup_{z\in\Z} T_z\Z$, with latent controller variable $z \in \Z \subseteq \mathbb{R}^{\nz}$:
\begin{subequations}\label{eqn:latent}\vspace{-33pt}
    \begin{multicols}{2}
        \begin{equation}\label{eq:controller}
            u = k(z, y_k),
        \end{equation}\break
        \begin{equation}\label{eq:latent_dyn}
            \dot z = l(z, y_k),
        \end{equation}
    \end{multicols}\vspace{-8pt}
\end{subequations}
\noindent and which when applied to system \eqref{eq:system}, maximizes the volume of the closed-loop ROA around a desired equilibrium point $x_0$. To make the synthesis and verification of this controller compatible with SOS programming, we search over a subset of \textit{polynomial} controllers by parameterizing $k$ and $l$ as a polynomial of $z$ and $y_k$, i.e., $u = \theta_k^\top m_k(z, y_k)$ and $l = \theta_l^\top m_l(z, y_k)$, for monomial bases $m_{k/l}(z, y_k)$ of degree $d_k$ and $d_l$, respectively. Note that while \eqref{eq:latent_dyn} is not explicitly a function of $u$ (to avoid bilinearities between $\theta_c$ and $\theta_m$), \eqref{eq:latent_dyn} can still recover $u$ since \eqref{eq:controller} is also a polynomial of $y_k$ and $z$. 
To find an inner approximation of the ROA, we first define the augmented state $\aug = [x^\top, z^\top]^\top \in \X \times \Z \doteq \A$ and dynamics $\dot \aug = [f(x, u)^\top, l(z,y_k)^\top]^\top$. Then, ideally, we wish to jointly search for a Lyapunov function $V: \X \times \Z \rightarrow \mathbb{R}_{\ge 0}$, controller $k$, and latent dynamics $l$ that solves
\begin{equation}\label{eq:ideal}%\small
    \begin{array}{cl}
       \underset{V, k, l}{\textrm{maximize}}  & \textrm{Vol}(\Omega_\rho) \\
        \textrm{subject to} & V(x,z) > 0,\quad \forall (x, z) \ne (x_0,0)\doteq \aug_0 \\ 
        & V(x_0,0) = 0 \\
        & \dot V(x,z) < 0,\quad \forall x \in \Omega_\rho,
    \end{array}
\end{equation}
where the $\rho$-sublevel set of $V$ is denoted $\Omega_\rho \doteq \{x,z \mid V(x,z) \le \rho\}$, and $\rho$ is fixed (to set the scaling of $V$). The controller returned by \eqref{eq:ideal} is guaranteed to stabilize any initial conditions (ICs) $(x, z) \in \Omega_\rho$ to the augmented goal $\aug_0$. While the pointwise (non)-negativity constraints can be handled via SOS, \eqref{eq:ideal} is still nonconvex, due to bilinearities between $V$ and $k$, $l$ in the final constraint of \eqref{eq:ideal}; moreover, additional nonconvexities arise with the constraints needed to model input constraints. Given the challenge of solving \eqref{eq:ideal} exactly, we propose methods to approximately solve \eqref{eq:ideal} in Sec. \ref{sec:method}.

\section{Control Synthesis and Verification}\label{sec:method}
\begin{figure}
    \centering
    \includegraphics[width=\linewidth]{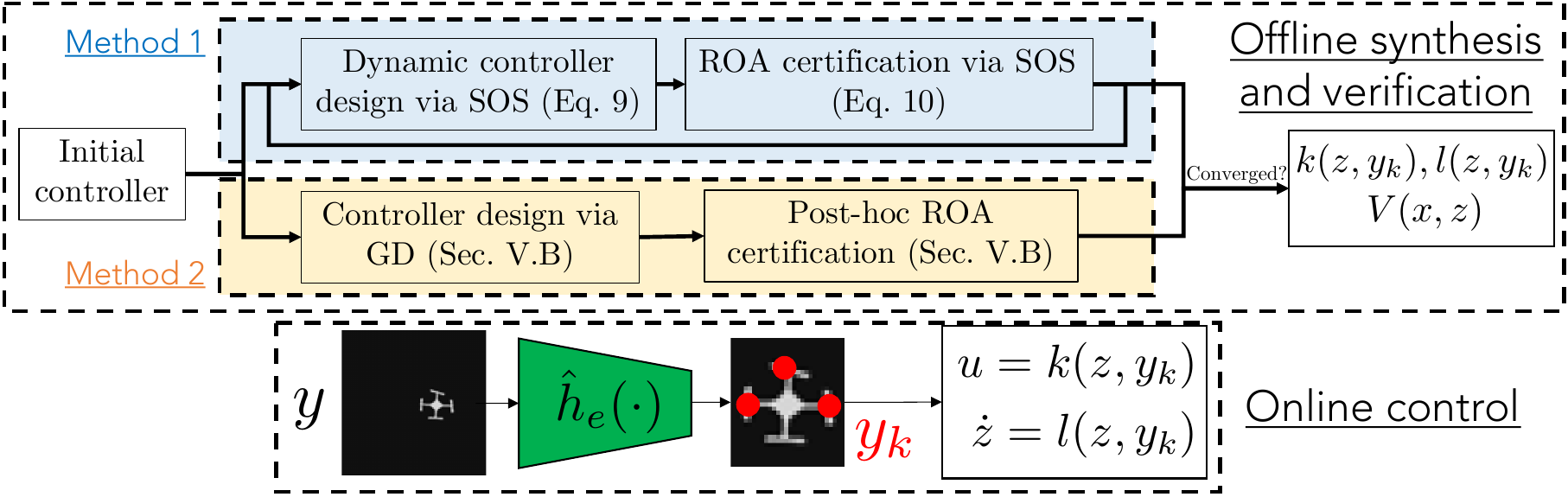}\vspace{-10pt}
    \caption{Method overview. \textbf{Offline}: we synthesize a dynamic-output-feedback control policy, either through SOS or through direct gradient-based optimization. \textbf{Online}: we compose the learned perception system with the output-feedback policy to obtain a control action from an input image.}
    \label{fig:method_flowchart}
    \vspace{-15pt}
\end{figure}

We propose two methods for approximately solving \eqref{eq:ideal} (see Fig. \ref{fig:method_flowchart} for an overview). We first apply bilinear alternations (Sec. \ref{sec:method_alternations}), i.e., we perform coordinate ascent on the ROA volume by fixing alternating subsets of the variables in \eqref{eq:ideal} and solving the resulting convex subproblems. Our second method directly learns $\theta_c$ and $\theta_m$ through GD (Sec. \ref{sec:method_gradient}) and certifies an ROA \textit{a posteriori}. We compare the two methods in Sec. \ref{sec:results}, and discuss their tradeoffs in Sec. \ref{sec:conclusion}.

\vspace{-5pt}
\subsection{Method 1: SOS alternations}\label{sec:method_alternations}

Our alternation scheme is summarized in Alg. \ref{alg:alternations}. We break down each individual optimization below.
\begin{figure}[!h]\vspace{-12pt}
 \removelatexerror
\begin{algorithm}[H]\label{alg:alternations}\small\DontPrintSemicolon
\KwIn{$\theta_k^\textrm{init}$, $\theta_l^\textrm{init}$ \tcp*{initial controller parameters\hspace{-15pt}}}
$V, L \leftarrow$ solve \eqref{eq:alternation_init1} using $(\theta_k^\textrm{init}, \theta_l^\textrm{init})$\\
$\rho, L_b \leftarrow$ solve \eqref{eq:alternation_init2} using $(V)$\\
$\hat E \leftarrow I$\\
\For{$j = 1, \ldots, \textrm{MAX ITER}$}{
    $\rho, \theta_k, \theta_l, E, L, L_\textrm{ell} \leftarrow$ solve \eqref{eq:alternation_1} using $(V, \rho, \hat E)$\\
    $E, V \leftarrow$ solve \eqref{eq:alternation_2} using $(\theta_k, \theta_l, L, L_\textrm{ell}, \rho, \hat E)$\\
    $\hat E \leftarrow E$\\
}
\Return $V, \rho, \theta_k, \theta_l$
\caption{Bilinear alternations for solving \eqref{eq:ideal}}
\end{algorithm}\vspace{-15pt}
\end{figure}

First, given a set of initial controller parameters $\theta_k^\textrm{init}$ and $\theta_l^\textrm{init}$ (see Sec. \ref{sec:method_init} on how we initialize), we search for a Lyapunov function valid in a ball around $\aug_0$ (Alg. \ref{alg:alternations}, line 1):
\begin{equation}\label{eq:alternation_init1}%\small
	\begin{array}{cl}
		\textrm{find} & V, L \\
		\textrm{subject to} & V \textrm{ is SOS}, \quad L \textrm{ is SOS} \\
		& -\frac{\partial V}{\partial \aug}^\top \dot \aug + L(\Vert \aug-\aug_0 \Vert_2^2 - r)\textrm{ is SOS}.
	\end{array}\hspace{-5pt}
\end{equation}
This enforces $\dot V < 0$ over a ball of radius $r$ centered at $\aug_0$. To find an initial ROA, we find the largest sublevel set of $V$, $\Omega_\rho$, contained in this ball, by solving for a fixed $r$ (line 2):

\vspace{-13pt}
\begin{equation}\label{eq:alternation_init2}%\small
	\begin{array}{cl}
		\underset{\rho, L_b}{\textrm{maximize}} & \rho \\
		\textrm{subject to} & V - \rho + L_b(r - \Vert \aug-\aug_0 \Vert_2^2) \textrm{ is SOS} \\
                                & L_b \textrm{ is SOS}.
	\end{array}	\hspace{-5pt}
\end{equation}

Then, in line 5, given the fixed Lyapunov function $V$ and sublevel set $\Omega_\rho$, we search for $\theta_k$, $\theta_l$ and SOS Lagrange multipliers $L$ for enforcing the Lyapunov conditions over $\Omega_\rho$. As a surrogate for maximizing the volume of $\Omega_\rho$, we maximize the volume of an ellipsoid $\mathcal{E}$ inscribed within $\Omega_\rho$: $\mathcal{E} \doteq \{\aug \mid (\aug - \aug_0)^\top E(\aug-\aug_0) \le 1\}$. This containment condition can be enforced by the following SOS constraints:

\vspace{-10pt}
\begin{subequations}\label{eq:ell}\vspace{-3pt}
	\begin{align}
			(\aug-\aug_0)^\top E(\aug-\aug_0) - 1 + L_\textrm{ell}(\rho - V) \textrm{ is SOS}, \label{eq:ell_a} \\
			L_\textrm{ell} \textrm{ is SOS}, \label{eq:ell_b}
	\end{align}
\end{subequations}
while maximizing the volume of $\mathcal{E}$ can be done by minimizing the log determinant of $E$. As $\log\det(E)$ is concave in $E$, its minimization is non-convex; thus, we minimize a linearization of $\log\det(\cdot)$ around the ellipsoid from the previous iteration $\hat E$, which can be written as $\log\det(E) \approx \log\det(\hat E) + \mathrm{tr}(\hat E^{-1} (E - \hat E))$ \cite{DBLP:conf/amcc/FazelHB03}. Removing constants in the linearization and putting everything together gives
\begin{equation}\label{eq:alternation_1}%\small
	\hspace{-1pt}\begin{array}{cl}
		\underset{\theta_k,\theta_l, E, L, L_\textrm{ell}}{\textrm{minimize}} & \mathrm{tr}(\hat E^{-1} E) \\
		\textrm{subject to} & -\frac{\partial V}{\partial \aug}^\top \dot \aug(\theta_k, \theta_l) + L(V - \rho) \textrm{ is SOS} \\
            & L \textrm{ is SOS},\quad \textrm{Eq. } \eqref{eq:ell_a},\quad \textrm{Eq. } \eqref{eq:ell_b} \\
	\end{array}\hspace{-9pt}
\end{equation}
However, note that $V$ and $\rho$ are fixed in \eqref{eq:alternation_1}, and the objective is just meant to provide a Lagrange multiplier $L_\textrm{ell}$ for the tightest inscribed ellipsoid in $\Omega_\rho$, which is to be used in the next alternation step \eqref{eq:alternation_2}. To explicitly increase the ROA in \eqref{eq:alternation_1}, we can maximize $\rho$ in an outer maximization via bisection search, which aims to find a controller that increases the ROA with respect to the current candidate $V$.

Finally, for fixed controller parameters and Lagrange multipliers, we aim to find an improved Lyapunov function $V$ that can certify a larger ROA for the current controller by maximizing the volume of an ellipsoid inscribed in $\Omega_\rho$:
\begin{equation}\label{eq:alternation_2}
	\begin{array}{cl}
		\underset{E, V}{\textrm{minimize}} & \mathrm{tr}(\hat E^{-1} E) \\
		\textrm{subject to} & -\frac{\partial V}{\partial \aug}^\top \dot \aug + L(V - \rho) \textrm{ is SOS} \\
		& \quad L \textrm{ is SOS},\quad \textrm{Eq. } \eqref{eq:ell_a}
	\end{array}\hspace{-9pt}
\end{equation}

\subsubsection{Initializing the alternations}\label{sec:method_init}
Alg. \ref{alg:alternations} requires an initial guess for the controller. In the full-state reconstruction case, i.e., $\nz = \nx$, we can solve LQG for the linearization of \eqref{eq:dyn}-\eqref{eq:kp} around $\aug_0$ to obtain a locally-stable initialization. However, for the reduced-order case, i.e., $\nz < \nx$, the separation principle does not hold and solving the reduced-order LQG problem involves nonlinear solvers \cite{doi:10.2514/6.1984-1035} which may not converge. As an alternative, we use GD to optimize sampled closed-loop trajectories to initialize the controller.

In particular, we initialize $\theta_k$ and $\theta_l$ to random values, sample ICs $\{x_i\}_{i=1}^{N_\textrm{samp}}$ near the equilibrium $\aug_0$, roll out the policy for a fixed time horizon $T$ on the $\Delta T$-time discretized dynamics \eqref{eq:dyn} to obtain $N_\textrm{samp}$ trajectories $\xi_i \doteq \{x_i^t, u_i^t\}_{t=1}^T$. We define a cost on trajectories $c(\xi) = c_T(x_T) + \sum_{t=1}^{T-1} \alpha_\aug \Vert \aug_t - \aug_0\Vert_2^2 + \alpha_u \Vert u_t \Vert_2^2$ for weighting parameters $\alpha_\aug, \alpha_u \ge 0$, and evaluate the cost of each trajectory $c(\xi_i)$. Finally, we define our policy loss as an averaged cost over trajectories,
\begin{equation}\label{eq:loss}
	\mathcal{L}\doteq \textstyle\frac{1}{N_\textrm{samp}} \textstyle\sum_{i=1}^{N_\textrm{samp}} c(\xi_i),
\end{equation}
and minimize \eqref{eq:loss} via automatic differentiation (e.g., in PyTorch) to improve $\theta_k$ and $\theta_l$. Minimizing \eqref{eq:loss} draws the closed-loop trajectories toward $\aug_0$, thereby increasing the ROA size. 
We now discuss extensions to Alg. \ref{alg:alternations}.
\subsubsection{Control input constraints}
We can ensure that the synthesized controller can stabilize the system even in the presence of control constraints through adding additional constraints. For example, for the scalar-valued input case, to ensure stabilization given an upper control limit $u \le \bar u$, we can enforce $\frac{\partial V}{\partial x}f(x, \bar u) + \frac{\partial V}{\partial z} l < 0$ for all states in $\Omega_\rho \setminus \{\aug_0\}$ where $u(x) \ge \bar u$. This involves additional SOS multipliers $L_k$ (see Sec. IV of \cite{DBLP:conf/icra/MajumdarAT13} for details) and bilinearities between $L_k$ and $k$. To avoid these bilinearities, we search for these multipliers $L_k$ in \eqref{eq:alternation_2} (Alg. \ref{alg:alternations}, line 6).

\subsubsection{Trigonometric terms}
To handle trigonometric terms which arise in rigid body dynamics, e.g., $\sin(\phi)$ for the angle $\phi$ of an inverted pendulum, we can perform a change of variables to render the dynamics polynomial. Specifically, we replace any instances of $\sin(\phi)$ and $\cos(\phi)$ in \eqref{eq:system} with auxiliary state variables $s$ and $c$, and add an additional constraint $s^2 + c^2 = 1$. As the dynamics are constrained, we only need to enforce the Lyapunov conditions over $\{x \mid s^2 + c^2 = 1\}$, which can be enforced with additional multipliers in each SOS program in Alg. \ref{alg:alternations}, i.e., $L_t (s^2 + c^2 - 1)$, where $L_t$ is a polynomial. As $L_t$ does not multiply with any decision variables, it does not complicate the alternation scheme.

\subsubsection{Observation error}
To model the impact of observation error in \eqref{eq:kp} on closed-loop stability, we can add additional indeterminates $w$ in the SOS program. For instance, given a uniformly bounded disturbance $\W = \Vert w \Vert_2 \le \bar w$, we can enforce the Lyapunov conditions to hold robustly for all $w \in \W$, e.g., the first constraint of \eqref{eq:alternation_1} becomes
\begin{equation}\label{eq:sos_obs}
	-\textstyle\frac{\partial V}{\partial \aug}^\top \dot \aug(\theta_k, \theta_l) + L(V - \rho) + L_w(\Vert w \Vert_2^2 - \bar w) \textrm{ is SOS}.
\end{equation}
for SOS multipliers $L_w$. In general, our formulation can handle an observation error description written as a set of polynomial (in)-equalities in $x$, i.e., $\W$ is a basic semialgebraic set. In this paper, when controlling from images, we bound our keypoint extractor error using a uniform error bound $\Vert w \Vert_2 \le \bar w$ valid over a set $\A_w \subseteq \A$. To over-estimate $\bar w$ with high probability, we can leverage extreme value theory, which estimates $\bar w$ from i.i.d. samples of the error $\Vert \hat h_e(h(x)) - h_k(x)\Vert $ for $x$ sampled from $\A_w$ by following the approach laid out in \cite{DBLP:journals/corr/abs-2212-06874, Coles2001, DBLP:journals/ral/KnuthCOB21, arxiv_version}.

\subsubsection{Implicit formulation}\label{sec:method_implicit}

For systems that can be modeled with rational polynomial dynamics (e.g., the cart-pole), it can be easier to determine the system's ROA by writing its dynamics in implicit form, i.e., as a set of polynomial equalities $g(\aug, u, \dot \aug) = 0$, which can eliminate rational terms that appear when the dynamics are written explicitly as in \eqref{eq:dyn}. To adapt the Lyapunov conditions \eqref{eq:lyap} to systems in implicit form, we can ensure that $\dot V < 0$ via additional indeterminates $b$ and enforcing $-\frac{\partial V}{\partial \aug}b + L_g g(\aug, u, b) \textrm{ is SOS}$, where the Lagrange multipliers $L_g$ are polynomials.

Specifically, the alternation scheme in Alg. \ref{alg:alternations} is modified as follows. In line 1, we modify \eqref{eq:alternation_init1} to also search for $L_g$:
\begin{equation}\label{eq:alternation_init1_implicit}\small
		\hspace{-7pt}\begin{array}{cl}
		\textrm{find} & V, L, L_g \\
		\textrm{subject to} & V \textrm{ is SOS},\quad L \textrm{ is SOS} \\
		& -\frac{\partial V}{\partial \aug}^\top \dot \aug + L_g g + L(\Vert \aug-\aug_0 \Vert_2^2 - r) \textrm{ is SOS}.
	\end{array}\hspace{-12pt}
\end{equation} 
Lines 2-4 of Alg. \ref{alg:alternations} remain the same. We replace line 5 of Alg. \ref{alg:alternations} with the following optimization which finds Lagrange multipliers and the inscribed ellipsoid $\mathcal{E}$
\begin{equation}\label{eq:alternation_1_implicit}%\small
	\hspace{-1pt}\begin{array}{cl}
		\underset{E, L, L_\textrm{ell}, L_g}{\textrm{minimize}} & \mathrm{tr}(\hat E^{-1} E) \\
		\textrm{subject to} & -\frac{\partial V}{\partial \aug}^\top b + L_g g + L(V - \rho) \textrm{ is SOS} \\
		& L \textrm{ is SOS},\quad \textrm{Eq. } \eqref{eq:ell}.
	\end{array}\hspace{-9pt}
\end{equation}
Finally, we replace line 6 of Alg. \ref{alg:alternations} with a simultaneous search for $V, \theta_k, \theta_l, E, L_\textrm{ell}$.
\begin{equation}\label{eq:alternation_2_implicit}\small
	\hspace{-1pt}\begin{array}{cl}
		\underset{E, V, \theta_k, \theta_l, L_\textrm{ell}}{\textrm{minimize}} & \mathrm{tr}(\hat E^{-1} E) \\
		\textrm{subject to} & -\frac{\partial V}{\partial \aug}^\top b + L_g g(\theta_k, \theta_l) + L(V - \rho) \textrm{ is SOS} \\
		& \textrm{Eq. } \eqref{eq:ell_a}
	\end{array}\hspace{-9pt}
\end{equation}

\subsection{Method 2: Gradient-based synthesis}\label{sec:method_gradient}

As an alternative to running control design alternations as in Sec. \ref{sec:method_alternations}, we can opt to directly learn the dynamic controller parameters $\theta_k$ and $\theta_l$ through minimizing \eqref{eq:loss}. That is, instead of using the gradient-based approach of Sec. \ref{sec:method_init} just to obtain an initialization for Alg. \ref{alg:alternations}, we can select a larger set $\A_s$ over which we want the closed-loop system to be stable, sample ICs from $\A_s$, and exactly follow the procedure in Sec. \ref{sec:method_init}. Given a set of learned parameters $\theta_k$ and $\theta_l$, we can find an inner approximation of the ROA of the closed-loop system by following the alternation scheme of Alg. \ref{alg:alternations}, but with the simplification that $\theta_k$ and $\theta_l$ are fixed. This is an advantage over learning an NN policy, which are more difficult to verify than our simple policies.
Overall, compared to alternations (Alg. \ref{alg:alternations}), this method requires more parameter tuning, i.e., of the horizon $T$ and time-step $\Delta T$, in order for the policy learning to reliably converge. However, for some systems, the policies obtained using GD can have a larger ROA than what is obtained with alternations (see Sec. \ref{sec:results} for comparisons with Alg. \ref{alg:alternations} and Sec. \ref{sec:conclusion} for discussion).

\vspace{-5pt}
\section{Results}\label{sec:results}
\vspace{-5pt}

We evaluate our method on an inverted pendulum, cart-pole, and planar/3D quadrotors. We aim to show that our method provides controllers with large ROAs that outperform 1) popular RL-based methods for output-feedback, while remaining simpler and easily verifiable, and 2) the more common approach of output-feedback through separate controller and estimator synthesis. Thus, we compare with 1) PPO \cite{rl-zoo3} using a recurrent neural network (RNN) policy (to handle partial observability) and given \textit{perfect, noiseless} keypoint observations (i.e., images are \textit{not} given as input) trained until convergence, and 2) full-order LQG for the linearization around the goal, which uses the separation principle \cite{astrom} to independently synthesize a locally-stable full-state-feedback controller and state estimator, returning state estimate $\hat x$. Our SOS programs are implemented with SumOfSquares.jl \cite{legat2017sos}, and the SOS constraints are interpreted with the Chebyshev basis to improve the numerical stability \cite{10.5555/2430708} of Alg. \ref{alg:alternations}. Images for the quadrotor examples were rendered with PyBullet.

\begin{figure}
    \centering
    \includegraphics[width=0.95\linewidth]{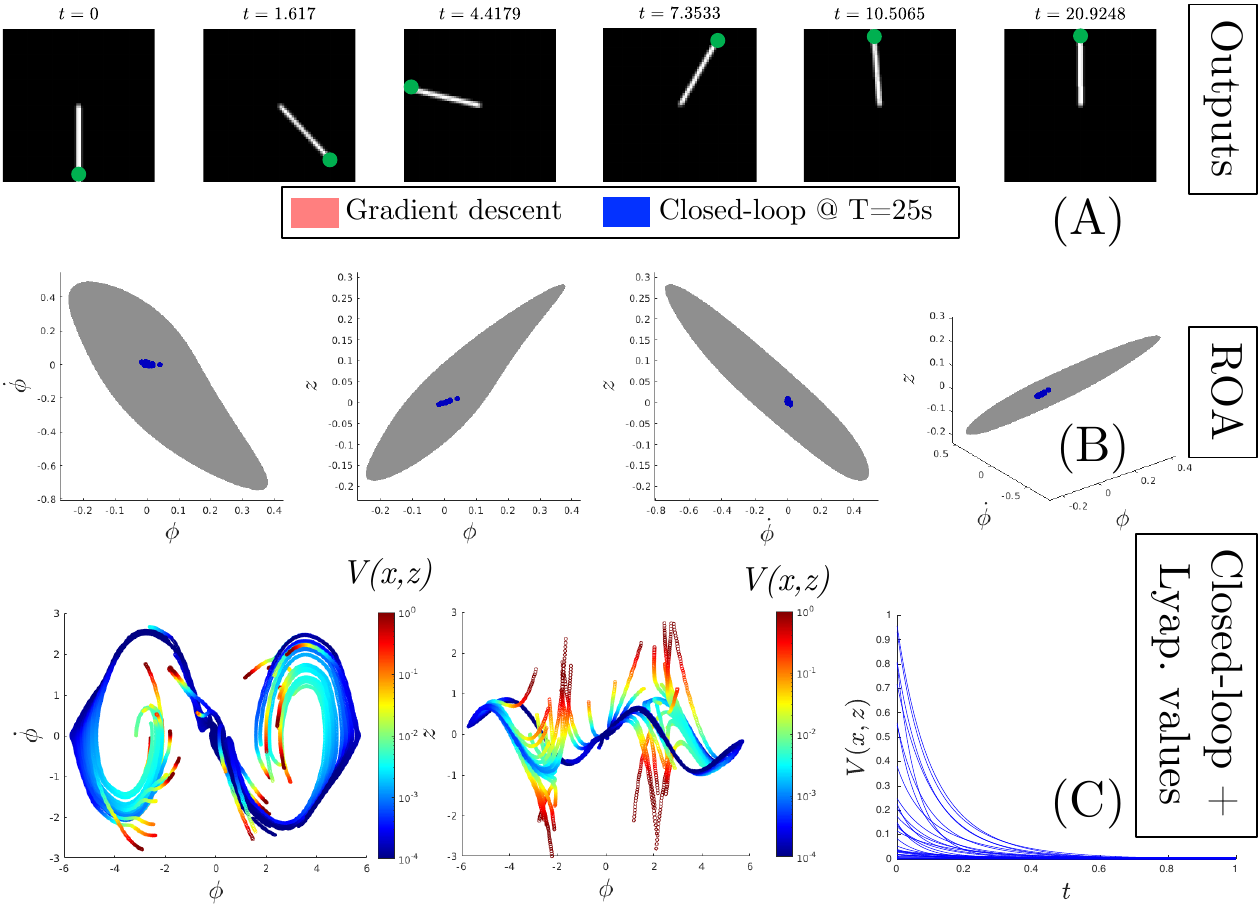}\vspace{-12pt}
    \caption{Stabilizing the inverted pendulum from images, using learned keypoints $y_k$ and a controller from \textit{Method 2}. \textbf{(A)}: Images received when stabilizing from $\aug = [\phi, \dot\phi, z]^\top = [\pi, 0, 0]^\top$; $y_k^*$ are marked in green. \textbf{(B)}: left to right, \textit{Projections} of invariant set ($\Omega_{4e-5}$) (gray) onto the $\phi\dot\phi$, $\phi z$, $\phi\dot z$ axes, and the full 3D invariant set. Here, $\Omega_{4e-5}$ is a sublevel set of a $V$ which certifies global convergence to $\Omega_{4e-5}$ under observation error caused by $\hat h_e$; for 150 randomly sampled ICs, we plot the closed-loop states reached after $25$s (blue). \textbf{(C)}: Closed-loop rollouts when controlling from images (left, center), color is $V$ value; $V$ along trajectories (right).}
    \label{fig:ip_roa}
    \vspace{-10pt}
\end{figure}

\begin{figure}
    \centering
    \includegraphics[width=0.9\linewidth]{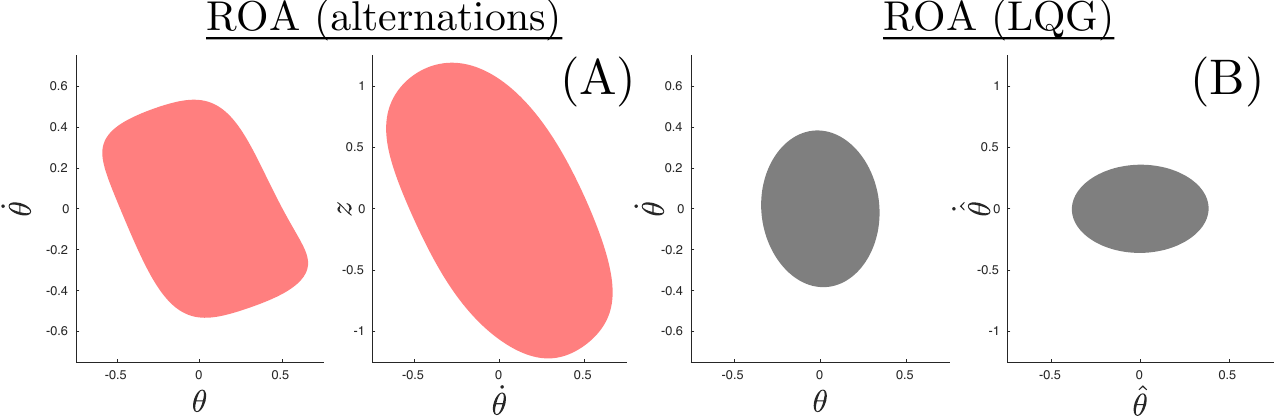}\vspace{-13pt}
    \caption{ROA slices for the inverted pendulum, using perfect keypoints $y_k^*$. \textbf{(A)}: ROA of controller from \textit{Method 1} (red). \textbf{(B)} ROA of LQG (gray).}
    \label{fig:ip_roa_alternations}
    \vspace{-18pt}
\end{figure}

\noindent\textbf{Inverted pendulum}: We consider the swing-up problem for a torque-limited inverted pendulum \cite[Ch. 2]{tedrake2009underactuated}. This system has state $x = [s, c, \dot\phi]^\top$, where $s$ and $c$ are the sine and cosine of the angular deviation $\phi$ from the upright equilibrium $x_0 = [0, 1, 0]^\top$, which we wish to stabilize to. We set the pendulum mass and length as $m = 1$kg and $\ell = 5$m, leading to a gravity torque $mg\ell$ of $49.05$ N$\cdot$m, while our torque limits are $25$ N$\cdot$m. Thus, we cannot directly overcome gravity to swing the pendulum to $x_0$, and must iteratively pump energy into the system to reach $x_0$. For outputs, we are given a single keypoint at the tip of the pendulum, i.e., we have the observation function $y_k^* = [-\ell s, \ell c]^\top \in \mathbb{R}^2$. We train $\hat h_e$, represented as a CNN, from a dataset of 4000 labeled pairs of 64x64 grayscale images and corresponding keypoints. We synthesize a degree 2 dynamic-output-feedback controller, i.e., $d_k = d_l = 2$, with a single latent state $\nz = 1$, using the GD strategy (Method 2) in Sec. \ref{sec:method_gradient}. When controlling using the perfect keypoints $y_k^*$, we prove that the synthesized policy has a global ROA to $\aug_0$ using a degree 6 Lyapunov function $V$. When using the learned keypoints $y_k = \hat{h}_e(y)$, we bound the perception error as $\Vert w \Vert_2 \le 0.003$, and certify global convergence to the $4\cdot 10^{-5}$-sublevel set of $V$, $\Omega_{4e-5}$, which we show in Fig. \ref{fig:ip_roa}(B) in gray. This set is invariant, since it is compact and satisfies $\dot V < 0$ on its boundary. To empirically show the invariance of $\Omega_{4e-5}$, we sample 150 ICs from $\Omega_{4e-5}^c$ and plot in Fig. \ref{fig:ip_roa}(B) (blue) the states reached by the closed-loop system after $25$s, which are all in $\Omega_{4e-5}$. We also show snapshots of the images used to stabilize from the downward-facing equilibrium in Fig. \ref{fig:ip_roa}(A), and example closed-loop rollouts (Fig. \ref{fig:ip_roa}(C)). We also synthesize a controller ($d_k = d_l = 3$, $\nz = 1$) via alternations (Method 1), which stabilizes near $\aug_0$ (see Fig. \ref{fig:ip_roa_alternations}, red) but cannot swing up from the downward equilibrium, as the alternations reach a local minimum in ROA volume. This is likely because the inscribed ellipsoid loosely approximates $\Omega_\rho$ for this system. 

For the baselines, we evaluate PPO by sampling 50 ICs from $[\phi, \dot\phi] \in [-10, 10]^2$ and computing the closed-loop 2-norm from $x_0$ after $60$s have elapsed. PPO swings up the pendulum to a neighborhood of $x_0$ for all 50 samples, achieving a goal error of $0.07 \pm 0.03$ (mean + stdev). Despite this, we note that the PPO policy does not render $x_0$ an equilibrium point (i.e., $u(x_0, z = 0) \ne 0$, where $z$ is the RNN hidden state), causing persistent chattering around $x_0$; thus, the goal error is nonzero. For the LQG baseline, we plot its certified ROA (obtained via SOS) in Fig. \ref{fig:ip_roa_alternations} (gray), which is smaller than the ROA achieved by both variants of our method. We note that for LQG to stabilize, the initial state estimate must be quite accurate; in contrast, our latent state $z$ is far less sensitive to initialization. Overall, these results suggest our method can effectively stabilize despite underactuation, that both variants of our method outperform LQG, and that our method is competitive with PPO.

\begin{figure}
    \centering
    \includegraphics[width=0.95\linewidth]{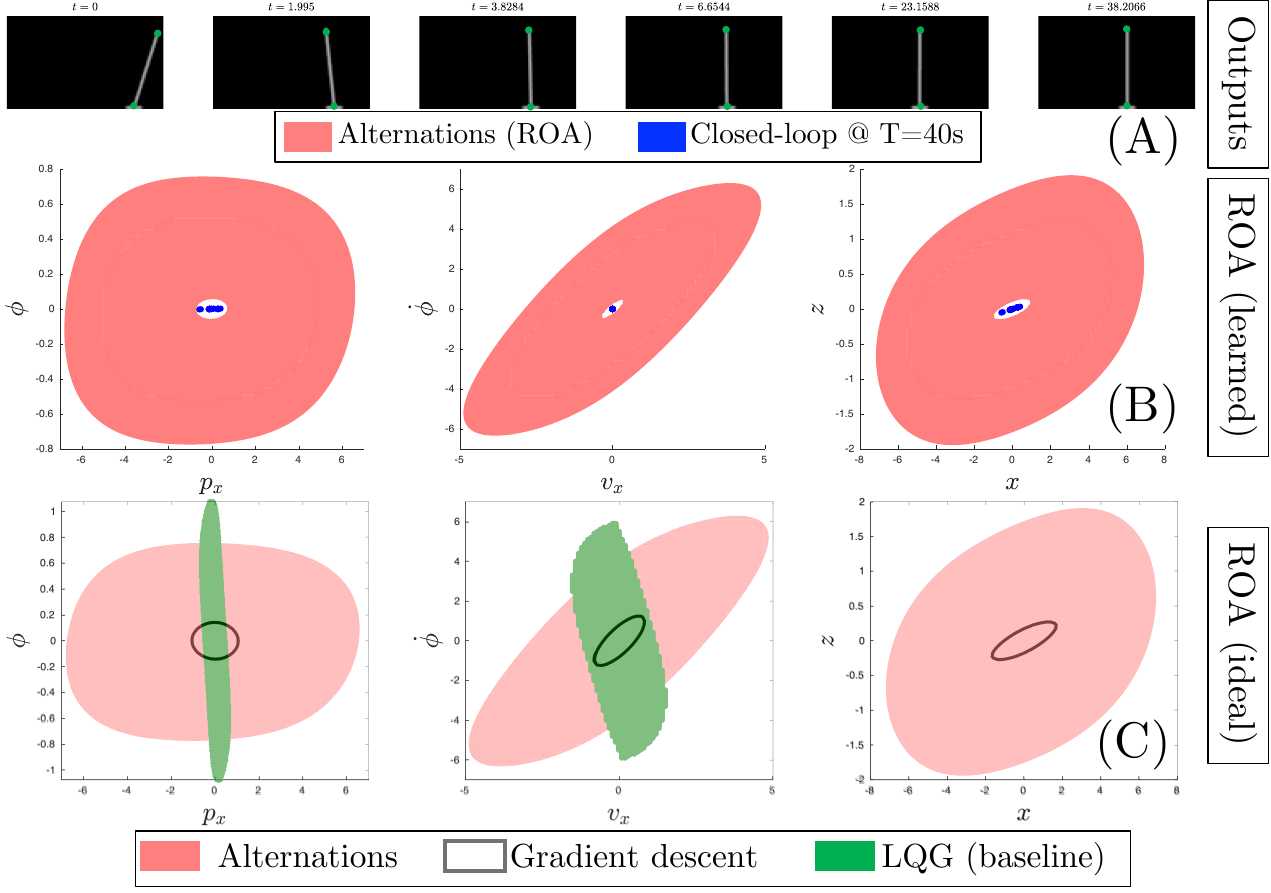}\vspace{-5pt}
    \caption{Stabilizing the cart-pole using learned keypoints $y_k$ \textbf{(A-B)} and perfect keypoints $y_k^*$ \textbf{(C)}. \textbf{(A)}: Snapshots of images used by our alternations-based controller to stabilize from $a_0 = [5.2, -0.3, 0.6, 0.6, 1.3, -0.3]^\top$; $y_k^*$ marked in green. \textbf{(B)}: ROA slice of our alternations-based controller when using $y_k$. \textbf{(C)}: ROA slices when using $y_k^*$. In red: Method 1 (alternations); black outline: Method 2 (gradient-based); in green: LQG.\vspace{-5pt} }
    \label{fig:cartpole_roa}
    \vspace{-16pt}
\end{figure}

\noindent \textbf{Cart-pole}: To show our approach can control systems with rational nonlinear dynamics, we synthesize a stabilizing policy for a cart-pole \cite[Ch. 3]{tedrake2009underactuated} This system has state $x = [p_x, s, c, \dot{p}_x, \dot\phi]^\top \in \mathbb{R}^5$. We wish to stabilize the system around the upright equilibrium, i.e., $x_0 = [0, 0, 1, 0, 0]^\top$. Using our alternations-based approach (Method 1), we synthesize a controller, where $d_k = 4$, $d_l = 1$, and $\nz = 1$. To avoid rational terms in the explicit dynamics, we use the implicit SOS variant of Alg. \ref{alg:alternations} discussed in Sec. \ref{sec:method_implicit}. For outputs, we are given two keypoints, one at the base and one at the tip of the pole, i.e., $y_k^* = [p_x, 0, p_x - \ell s, \ell c]^\top \in \mathbb{R}^4$ (see Fig. \ref{fig:keypoints}). We train $\hat h$, represented as a CNN, using 20000 pairs of labeled 56x96 grayscale images and keypoints, and bound their error as $\Vert w \Vert \le 0.05$, for all $\aug \in \Omega_{2.75}$. Under this error, we can certify using a degree 4 Lyapunov function that $\Omega_{2.75} \setminus \Omega_{0.01}$ converges to $\Omega_{0.01}$ (shown in Fig. \ref{fig:cartpole_roa}(B), white), and $\Omega_{0.01}$ is an invariant set. To show this invariance empirically, we plot the states reached after $40$s have elapsed in blue (Fig. \ref{fig:cartpole_roa}(B)) and images seen when stabilizing from a state in $\Omega_{2.75}\setminus\Omega_{0.01}$ in Fig. \ref{fig:cartpole_roa}(A). When using perfect keypoint observations $y_k^*$, we can certify that the entirety of $\Omega_{2.75}$ is contained in the ROA (Fig. \ref{fig:cartpole_roa}(C)). We also evaluate our GD variant (Method 2) with $d_k = 4$, $d_l = 1$, and $\nz = 1$; however, we cannot effectively descend on \eqref{eq:loss}, yielding a controller with a small ROA (Fig. \ref{fig:cartpole_roa}(C), black). 

In terms of baselines, for PPO, we attempt to learn a stabilizing policy over $[p_x, \phi, \dot{p}_x, \dot\phi] \in [-0.5, 0.5]^4$. While the PPO policy can reliably maintain the pole's orientation near $\phi = 0$, the closed-loop cart position $p_x$ continuously oscillates around zero and fails to stabilize to $x_0$ for all ICs, leading to large goal errors $6.1 \pm 0.8$ after $40$s have elapsed (averaged over 50 rollouts). For LQG, we plot states which empirically converge in closed-loop to $x_0$ (Fig. \ref{fig:cartpole_roa}(C), green), since computing the ROA using SOS is prohibitive (due to there being 20 indeterminates: 5 original states $x$, 5 implicit variables $b$, and 10 for the estimates of those variables $\hat x$, $\hat b$). While the ROA of LQG is larger in the $\phi$ dimension, it overall has smaller volume compared to our alternations-based controller, which can overcome much larger perturbations to the cart position $p_x$ and velocity $\dot p_x$. Overall, this experiment suggests that our model-based alternations approach can synthesize stronger controllers than the baselines for rational systems, and that alternations may more robustly improve the controller compared to gradient-based methods when the landscape of \eqref{eq:loss} is poorly-shaped.

\begin{figure}
    \centering
    \includegraphics[width=0.95\linewidth]{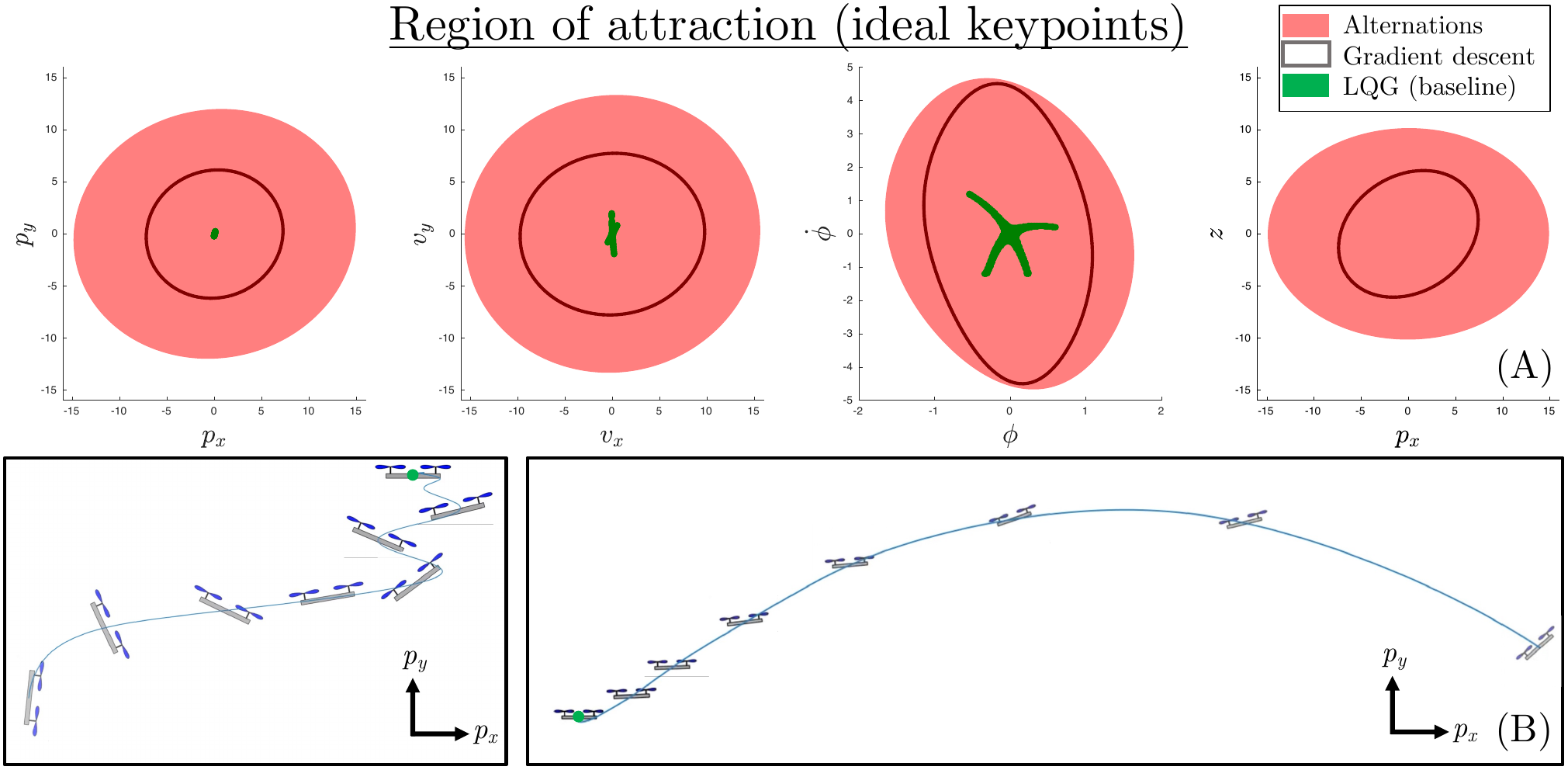}\vspace{-5pt}
    \caption{Stabilizing the planar quadrotor with \textit{perfect} keypoints $y_k^*$. \textbf{(A)}: ROA slices for a controller obtained via alternations (red), via GD (black), and sampled ICs in the ROA for LQG (green). \textbf{(B)}: Time-lapse of example trajectories of our closed-loop system, when controlling using the alternations-based policy driven by perfect keypoints.}
    \label{fig:pvtol_roa}
    \vspace{-20pt}
\end{figure}

\noindent\textbf{Planar quadrotor}: We demonstrate that our approach can stabilize a planar quadrotor from images. This system has state $x = [p_x, p_y, s, c, \dot{p}_x, \dot{p}_y, \dot\phi]^\top$, with dynamics as in \cite[Ch. 3]{tedrake2009underactuated}. We wish to stabilize to $x_0 = [0, 0, 0, 1, 0, 0, 0]^\top$ with input limits $u \in [0, 2mg]^2 \subseteq \mathbb{R}^2$, where $m=1$kg. We synthesize a linear dynamic controller, i.e., $d_k = d_l = 1$, with a single latent state $\nz = 1$, using alternations (Sec. \ref{sec:method_alternations}). For outputs, we have two keypoints, one at the center of the quadrotor, and one on the right propeller (see Fig. \ref{fig:keypoints}), i.e., $y_k^* = [p_x, p_y, p_x+\ell c, p_y+\ell s]^\top \in \mathbb{R}^4$. We train $\hat h_e$, represented as a CNN, from 40000 labeled pairs of 128x96 grayscale images and keypoints. When controlling with the ideal keypoints $y_k^*$, we can certify using a degree-2 Lyapunov function $V$ that $\Omega_{1.3}$ is an inner approximation of the controller's ROA (shown in Fig. \ref{fig:pvtol_roa}). We note that $\Omega_{1.3}$ contains states as distant as $15$m from the origin and orientations beyond $\pi/2$ (Fig. \ref{fig:pvtol_roa}, B). We also evaluate our GD approach (Method 2) with $d_k = d_l = \nz = 1$, which also achieves a large ROA (Fig. \ref{fig:pvtol_roa}(A), black), though smaller than that which is achieved by alternations. This is because we run into difficulties in descending on \eqref{eq:loss} when expanding the ROA to distant states, due to the degradation of the landscape of \eqref{eq:loss} for long-horizon trajectories (see Sec. \ref{sec:conclusion}).

When controlling from images using $\hat h_e$, we reuse the same $V$ and controller obtained from alternations, and aim to certify a smaller sublevel set of $V$, $\Omega_{0.4}$, due to the challenge of training an accurate, high-resolution keypoint extractor over a large range of states from a low-resolution image. We bound the error in the learned keypoints $\hat{h}_e(y)$ as $\Vert w \Vert \le 0.003$ for all $\aug \in \Omega_{0.4}$, and certify global convergence of ICs in $\Omega_{0.4}\setminus \Omega_{0.0035}$ to $\Omega_{0.0035}$, which is an invariant set, and is plotted in Fig. \ref{fig:pvtol_rollout}(C), white). To empirically show the invariance of $\Omega_{0.0035}$, we plot in Fig. \ref{fig:pvtol_rollout}(C) (blue) the states reached after rolling out the policy for $10$s, showing that states indeed reach and remain in $\Omega_{0.0035}$. We also plot example rollouts in Fig. \ref{fig:pvtol_rollout}(A-B) and snapshots of the images used to stabilize the system in Fig. \ref{fig:pvtol_rollout}(A-B) (left).  

In contrast, PPO is unable to learn a stabilizing policy to $x_0$ over $\Omega_{1.3}$, leading to a goal error of $9.9 \pm 2.9$ over 25 sampled ICs. We believe this is because the controls taken by PPO rapidly destabilize the quadrotor, providing poor signal in improving the controller; reward or observation clipping could possibly improve performance. The LQG controller also has poor performance (see Fig. \ref{fig:pvtol_roa}, green), for ICs where the closed-loop system is stable), and is particularly sensitive to incorrect position estimates. Like before, the ROA of LQG is prohibitive to compute using SOS due to the doubling of the number of states required by full-order state estimation. Overall, this experiment suggests that both variants of our method can yield stronger controllers than the baselines.

\begin{figure}
    \centering
    \includegraphics[width=0.92\linewidth]{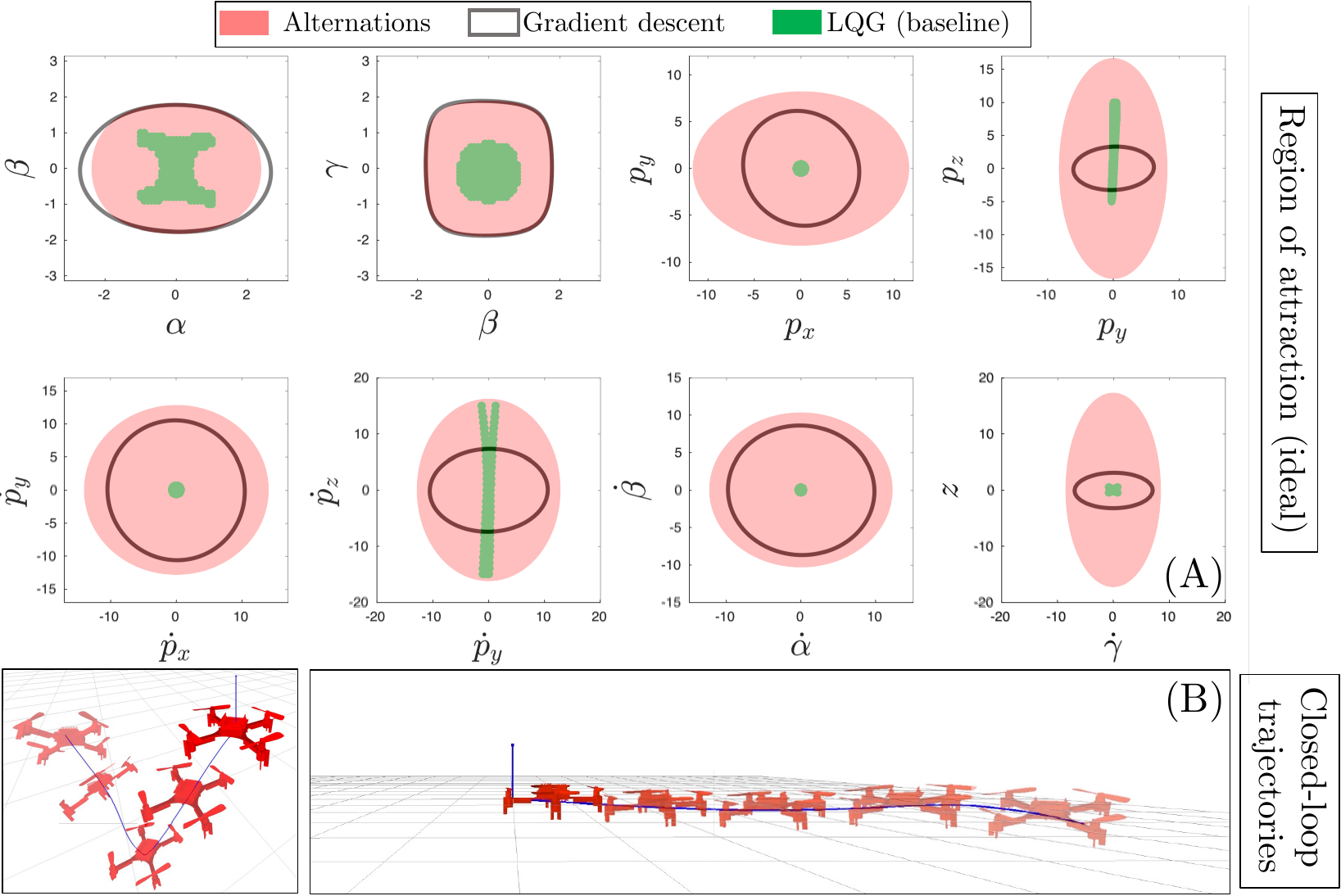}\vspace{-10pt}
    \caption{Stabilizing the 3D quadrotor from \textit{perfect} keypoints $y_k^*$. \textbf{(A)}: ROA slices for controller from alternations (red), from GD (black), and sampled ICs in the ROA for LQG (green). \textbf{(B)}: Time-lapse of closed-loop trajectories, when controlling using the alternations-based policy driven by $y_k^*$.}
    \label{fig:quad_roa}
    \vspace{-21pt}
\end{figure}

\begin{figure}
    \centering
    \includegraphics[width=0.95\linewidth]{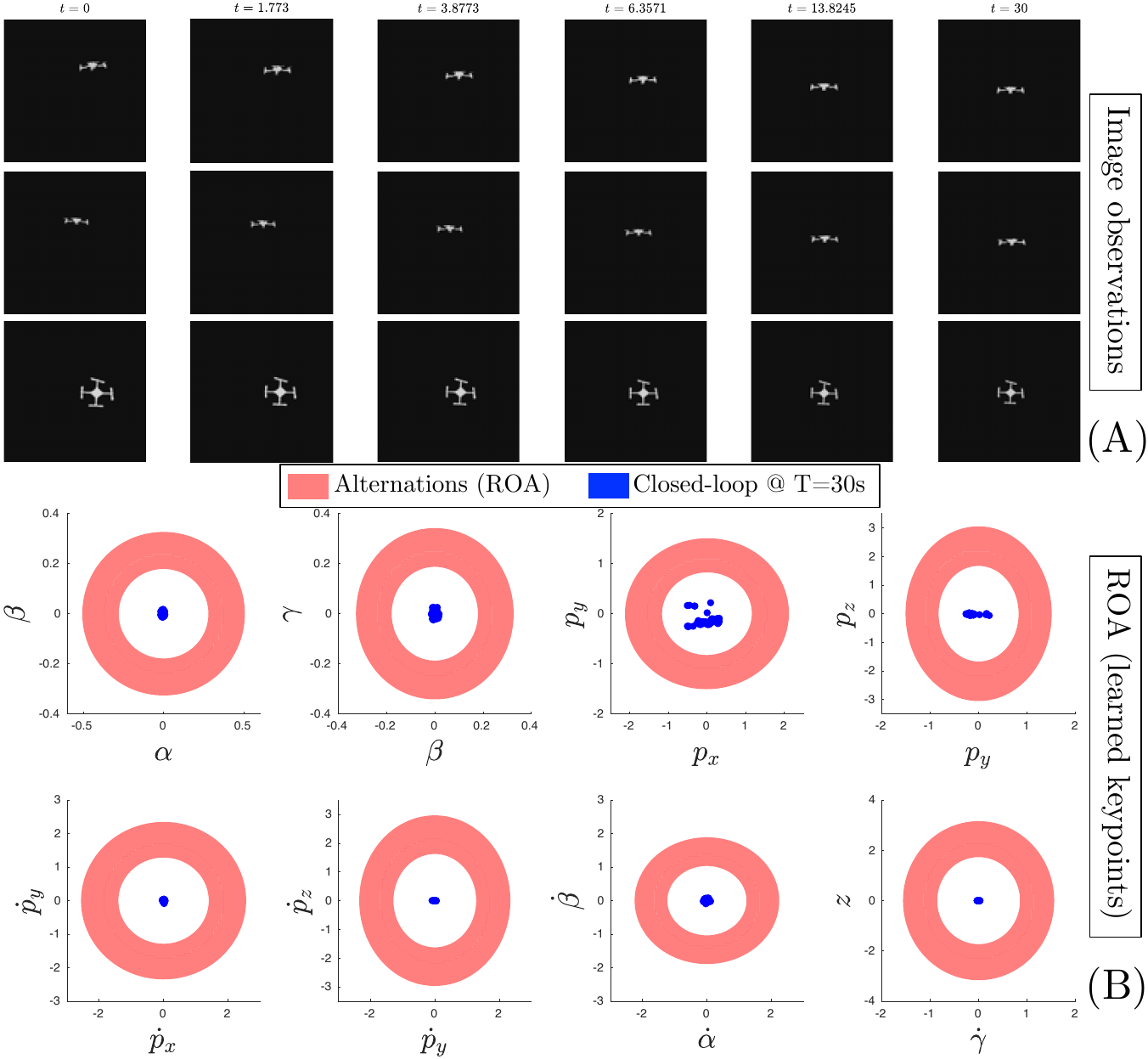}\vspace{-10pt}
    \caption{Stabilizing the 3D quadrotor from \textit{learned} keypoints $y_k$, using our alternations-based controller. \textbf{(A)}: snapshots of the three-view 128x128x3 images provided to the controller. \textbf{(B)}: verified ROA under keypoint error (red), which is guaranteed to converge to the white invariant set. We also plot (blue) the states reached in closed-loop, to show empirical invariance.}
    \label{fig:quad_roa_learned}
    \vspace{-22pt}
\end{figure}

\noindent \textbf{3D quadrotor}: Finally, to demonstrate our scalability to higher-dimensional systems, we synthesize a stabilizing visuomotor policy for a full 3D quadrotor. This system has state $x = [q_w, q_x, q_y, q_z, p_x, p_y, p_z, \dot{p}_x, \dot{p}_y, \dot{p}_z, \dot\alpha, \dot\beta, \dot\gamma]^\top \in \mathbb{R}^{13}$, where the quaternion $q_{(\cdot)}$-represented dynamics are given in \cite{fresk2013full} (we enforce the unit quaternion constraint via the S-procedure). We wish to stabilize to the origin $x_0 = [1,\mathbf{0}_{12}]^\top$, with input limits $u \in [0, 2.5mg/4]^4 \subseteq \mathbb{R}^4$. We synthesize a linear dynamic controller, i.e., $d_k = d_l = 1$, with a single latent state $\nz = 1$, using alternations (Method 1). For outputs, we have three keypoints, one on three of the four propellers (see Fig. \ref{fig:keypoints}), i.e., $y_k \in \mathbb{R}^9$. We train $\hat h_e$, represented as a CNN, from 150000 labeled pairs of 128x128x3 depth images and keypoints. Here, we stack three depth images, each recorded at different angles, offering views of the $p_xp_z$, $p_xp_y$, and $p_yp_z$ planes. When controlling with the ideal keypoints $y_k^*$, we certify using a degree-2 Lyapunov function $V$ that $\Omega_{0.3}$ is contained in the controller's ROA (shown in Fig. \ref{fig:quad_roa}). We note that $\Omega_{0.3}$ contains states as distant as $15$m from the origin and yaw/pitch/roll angles well beyond $\pi/2$ (Fig. \ref{fig:pvtol_roa}, B). We also evaluate GD (Method 2) with $d_k = d_l = \nz = 1$, which also achieves a large ROA (Fig. \ref{fig:quad_roa}(A), black), which while overall smaller in volume than the alternations ROA, has a larger ROA in the orientation states.

When controlling from images using $\hat h_e$, we reuse the same $V$ and controller from Method 1, and aim to certify a smaller sublevel set of $V$, $\Omega_{0.01}$ (shown in Fig. \ref{fig:quad_roa_learned}(B)), for the same reasons as for the planar quadrotor. We bound the error in the learned keypoints $\hat{h}_e(y)$ as $\Vert w \Vert \le 0.003$ for all $\aug \in \Omega_{0.01}$, and certify global convergence of ICs in $\Omega_{0.01}\setminus \Omega_{0.003}$ to $\Omega_{0.003}$, which is an invariant set, and is plotted in Fig. \ref{fig:quad_roa_learned}(B), white). To empirically show the invariance of $\Omega_{0.003}$, we plot in Fig. \ref{fig:quad_roa_learned}(B) (blue) the states reached after rolling out the policy for $30$s, showing that states indeed reach and remain in $\Omega_{0.003}$. We plot some images received along an example stabilization in Fig. \ref{fig:quad_roa_learned}(A).

In contrast, PPO is unable to learn a stabilizing policy to $x_0$ for ICs sampled from $\Omega_{0.01}$, leading to a goal error of $8.6 \pm 1.3$ over 25 ICs; we believe it fails for similar reasons as the planar quadrotor. The LQG controller also has poor performance (see Fig. \ref{fig:quad_roa}, green, where we plot sampled ICs where the closed-loop system is stable), and is highly sensitive to incorrect estimates in several state dimensions. Like before, the ROA of LQG is prohibitive to compute using SOS due to the doubling of states caused by full-order state estimation. Overall, this experiment suggests that both variants of our method can provide controllers with large ROA, even for high-dimensional systems, and provides huge computational savings compared to synthesizing/certifying a full-dimensional state estimator together with the controller.

\section{Discussion and Conclusion}\label{sec:conclusion}

We present two methods for synthesizing provably-stable reduced-order dynamic-output-feedback controllers from images: the first synthesizes stable-by-construction policies via bilinear alternations, and the second optimizes the controller via GD and certifies an ROA \textit{a posteriori}. Our method stabilizes several systems more effectively than baselines.

\noindent\textbf{Pros/cons of Method 1}: 
Alternation provides several benefits. For the full-order case (i.e., $\nz = \nx$), if the linearization around the goal is stabilizable and detectable, we can reliably initialize Alg. \ref{alg:alternations} with the controller/observer from LQG. Moreover, for both the full- and reduced-order cases, Alg. \ref{alg:alternations} monotonically increases the (ellipsoidal approximation of the) ROA volume. This systematic controller improvement makes Method 1 well-suited for expanding the ROA to distant regions of $\A$. The scalability of Method 1 is also similar to that of state-feedback synthesis for the systems in Sec. \ref{sec:results}, since a scalar $z$ is sufficient for stabilization. The biggest drawback of Method 1 is the numerical instability of Alg. \ref{alg:alternations}, though we find that orthogonal polynomial bases (like Chebyshev) greatly improve the numerics. Method 1 is also prone to local minima, especially when there is a large gap between the volume of $\Omega_\rho$ and its inscribed ellipsoid, i.e., for the inverted pendulum, which has a ``spiral"-shaped ROA (Fig. \ref{fig:ip_roa}(C)) that is poorly approximated by an ellipsoid.

\noindent\textbf{Pros/cons of Method 2}: 
GD is more scalable than SOS in high dimensions, though we may still need to solve a large SOS program to verify the controller. Method 2 also sidesteps local minima caused by the ellipsoid volume objective in Alg. \ref{alg:alternations}. Overall, GD is quite reliable for obtaining an initial controller with a small ROA around the goal. For drawbacks, Method 2 may not find a controller with certifiable ROA (as verification is done post-hoc), though we find empirically that closed-loop stability on sampled ICs generalizes well to stability on nearby ICs, due to the simplicity of our controller. There are also many parameters which must be tuned for success. Moreover, the landscape of \eqref{eq:loss} often has high Lipschitz constant, especially when $T$ is large \cite{DBLP:conf/iclr/0028MNRMGM22}, leading to myopic gradients that do not effectively descend \eqref{eq:loss}. This makes it difficult to expand the ROA to distant parts of $\A$ (as in the cart-pole and quadrotor examples). We believe these issues are not tied to the polynomial policy parameterization, as we also tried GD on an NN policy and observed the same issues. These issues can be mitigated by damping the system or using a critic to reduce the effective horizon required \cite{DBLP:conf/iclr/0028MNRMGM22}.

\noindent\textbf{Limitations and future work}: Though reducing the controller order helps to keep synthesis tractable, our approach, like state-feedback SOS methods, will still struggle to scale when the state dimension $n_x$ is high; thus, we will explore scalable variants like (S)DSOS \cite{DBLP:journals/siaga/AhmadiM19}. We also require labeled images and keypoints to train $\hat h_e$; to remove this limitation, we will explore unsupervised learning of polynomial latent dynamics from images, which would be amenable to SOS-based synthesis and verification tools.
\vspace{-8pt}

\bibliographystyle{IEEEtran}
\bibliography{references.bib}

\end{document}